\documentclass[runningheads]{llncs}
\usepackage{graphicx}
\usepackage{amssymb}
\usepackage{amsmath}
\usepackage{latexsym}
\usepackage{mathrsfs}
\usepackage{microtype}
\usepackage{subfigure}
\usepackage{booktabs} % for professional tables
\usepackage{caption}
\usepackage[utf8]{inputenc}
\usepackage{diagbox}
\usepackage{hyperref}
\usepackage{multirow}

\usepackage{listings}
\usepackage{color}

\definecolor{mygreen}{rgb}{0,0.6,0}
\definecolor{mymauve}{rgb}{0.58,0,0.82}
\definecolor{mygray}{rgb}{0.9,0.9,0.9}
\definecolor{codegreen}{rgb}{0,0.6,0}
\definecolor{codegray}{rgb}{0.5,0.5,0.5}
\definecolor{codepurple}{rgb}{0.58,0,0.82}
\definecolor{backcolour}{rgb}{0.95,0.95,0.92}
\definecolor{deepblue}{rgb}{0,0,0.5}
\definecolor{deepred}{rgb}{0.6,0,0}
\definecolor{deepgreen}{rgb}{0,0.5,0}

\lstdefinelanguage{Pythontf}{
	keywords= {False, class,        finally,        is,     return,
		None,   continue,       for,    lambda, try,
		True,   def,    from,   nonlocal,       while,
		and,    del,    global, not,    with,
		as,     elif,   if,     or,     yield,
		assert  ,else,  import  pass     ,
		break,  except  in,     raise, None, import, self},
	keywords=[2] {},
	morecomment=[l]{\#},
	morecomment=[s]{'''},
	morestring=[b]',
	morestring=[b]",
}

\newcommand\pythonstyle{\lstset{
		language=Pythontf,
		basicstyle=\fontfamily{fvm}\selectfont,
		keywordstyle=\color{deepblue},
		keywordstyle=[2]\color{orange},
		frame=tb,                         % Any extra options here
		showstringspaces=false,           %
		backgroundcolor=\color{backcolour},   
		commentstyle=\color{codegreen},
		stringstyle=\color{codepurple},
		showtabs=false,
		breaklines=true,                 
		tabsize=2,
		literate={~} {$\sim$}{1},
	}}
	
	\lstnewenvironment{python}[1][]
	{
		\small
		\pythonstyle
		\lstset{
			basicstyle=\ttfamily,
			columns=fullflexible,
		}
	}
	{}
	
	\newcommand\py[1]{{\Colorbox{backcolour}{\pythonstyle\lstinline!#1!}}}
\begin{document}
\title{LYRICS: a General Interface Layer to Integrate Logic Inference and Deep Learning\thanks{This project has received funding from the European Union's Horizon 2020 research and innovation program under grant agreement No 825619.}}

%\email{g.marra@unifi.it, \{fgiannini,diligmic,marco\}@diism.unisi.it} \\
%\url{http://sailab.diism.unisi.it/ } }
%
\author{Giuseppe Marra\inst{1,2} \and
	Francesco Giannini\inst{2} \and \\
	Michelangelo Diligenti\inst{2} \and Marco Gori\inst{2}}
\authorrunning{G. Marra et al.}
\institute{Department of Information Engineering, University of Florence, ITALY
	\email{g.marra@unifi.it}\\	
	\and
	Department of Information Engineering and Mathematical Sciences, \\ University of Siena, ITALY \\ \email{\{fgiannini,diligmic,marco\}@diism.unisi.it}}
\maketitle              % typeset the header of the contribution
\begin{abstract}
	In spite of the amazing results obtained by deep learning in many applications, a real intelligent behavior of an agent acting in a complex environment is likely to require some kind of higher-level symbolic inference. Therefore, there is a clear need for the definition of a general and tight integration between low-level tasks, processing sensorial data that can be effectively elaborated using deep learning techniques, and the logic reasoning that allows humans to take decisions in complex environments.
	This paper presents LYRICS, a generic interface layer for AI, which is implemented in TersorFlow (TF). LYRICS provides an input language that allows to define arbitrary First Order Logic (FOL) background knowledge. The predicates and functions of the FOL knowledge can be bound to any TF computational graph, and the formulas are converted into a set of real-valued constraints, which participate to the overall optimization problem. This allows to learn the weights of the learners, under the constraints imposed by the prior knowledge.
	The framework is extremely general as it imposes no restrictions in terms of which models or knowledge can be integrated. In this paper, we show the generality of the approach showing some use cases of the presented language, including model checking, supervised learning and collective classification.
	\keywords{deep learning, prior knowledge injection, first order logic}
\end{abstract}

\section{Introduction}
\label{introduction}
The success of deep learning relies on the availability of a large amount of supervised training data. This prevents a wider application of machine learning in real world applications, where the collection of training data is often a slow and expensive process, requiring an extensive human intervention.
%On the other hand, this raises concerns on what learning really means: how much deep learning simply relies on remembering previously seen patterns? How robust is such a system in an adversarial or uncontrolled environment, where outliers are either unpredictable or even forged to fool the trained system?
The introduction of prior-knowledge into the learning process is a fundamental step in overcoming these limitations. First, it does not require the training process to induce the rules from the training set, therefore reducing the number of required training data. Secondly, the use of prior knowledge can be used to express the desired behavior of the learner on any input, providing better behavior guarantees in an adversarial or uncontrolled environment.

This paper presents LYRICS,
%(Learning Yourself Reasoning and Inference with ConstraintS)
a TensorFlow~\cite{abadi2016tensorflow} environment based on a declarative language for integrating prior knowledge into machine learning, which allows the full expressiveness of First Order Logic (FOL) to define the knowledge. LYRICS has its root in frameworks like Semantic Based Regularization (SBR)~\cite{diligenti2012bridging,diligenti2015semantic} built on top of Kernel Machines and Logic Tensor Networks (LTN)~\cite{serafini2016learning} that can be applied to neural networks. These frameworks transform the FOL clauses into a set of constraints that are jointly optimized during learning. However, LYRICS generalizes both approaches by allowing to enforce the prior knowledge transparently at training and test time and dropping the previous limitations regarding the form of the prior knowledge. 
SBR and LTN are also hard to extend beyond classical classification tasks, where they have been applied in previous works, because the lack of a declarative front-end. On the other hand, LYRICS define a declarative language, dropping the barrier to build models exploiting the available domain knowledge in any machine learning context.

In particular, any many-sorted first-order logical theory can be expressed in the framework, allowing to declare domains of different sort, with constants, predicates and functions.
LYRICS provides a very tight integration of learning and logic as any computational graph can be bound to a FOL predicate. This allows to constrain the learner both during training and inference. Since the framework is agnostic to the learners that are bound to the predicates, it can be used in a vast range of applications including classification, generative or adversarial ML, sequence to sequence learning, collective classification, etc. %We provide some examples of application in the following of the paper.

%%%%%%%%%%%%%%%%%%%%%%%%%%%%%%%%%%%%%%%%%%%%%%%%%%%
\subsection{Previous work}
\label{sec:previous_work}
In the past few years many authors tackled specific applications by integrating logic and learning. Minervini et al.~\cite{minervini2017adversarial} proposes to use prior knowledge to correct the inconsistencies of an adversarial learner. Their methodology is designed ah-hoc for the tackled task, and limited to Horn clauses.
A method to distill the knowledge in the weights of a learner is presented by Hu et al.~\cite{hu2016harnessing}, which is also based on a fuzzy generalization of FOL. However, the definition of the framework is limited to universally quantified formulas and to a small set of logic operators. % with no discussion about the different selections of T-norms.
Another line of research~\cite{rocktaschel2015injecting,demeester2016lifted} attempts at using logical background knowledge to improve the embeddings for Relation Extraction.
However, these works are also based on ad-hoc solutions that lack a common declarative mechanism that can be easily reused. They are all limited to a subset of FOL and they allow to injecting the knowledge at training time, with no guarantees that the output on the test set respect the knowledge. %On the other hand, LYRICS is a general tool that allows to inject any background knowledge written in FOL logic with functions, both at training and test time.

Markov Logic Networks (MLN)~\cite{richardson2006markov} and Probabilistic Soft Logic (PSL)~\cite{kimmig2012short,bach2015hinge} provide a generic AI interface layer for machine learning by implementing a probabilistic logic.
%, whose parameters must be trained to determine the strength of the available knowledge in a given universe. MLN and PSL with their corresponding implementations have received lots of attention but they are limited by
However, the integration with the underlying learning processes working on the low-level sensorial data is shallow: a low-level learner can be trained independently, then frozen and stacked with the AI layer providing a higher-level inference mechanism. The language proposed in this paper instead allows to directly improve the underlying learner, while also providing the higher-level integration with logic.
TensorLog~\cite{cohen2016tensorlog} is a more recent framework to integrate probabilistic logical reasoning with the deep-learning infrastructure of TF, however TensorLog is limited to reasoning and does not allow to optimize the learners while performing inference.
TensorFlow Distributions~\cite{dillon2017tensorflow} and Edward~\cite{tran2017deep} are also a related frameworks for integrating probability theory and deep learning. %, maintaining computational efficiency and numerical stability.
However, these frameworks focus on probability theory and not the representation of logic and reasoning.

% We need space...
% The paper is organized as follows. In Section~\ref{sec:about} we introduce the framework, describe its declarative nature and delineate how first--order logic (FOL) formulas can be converted into a learning model. Section~\ref{zoo} shows the generality of the framework by showing a wide range of toy examples and applications. Finally, some conclusions are drawn in Section~\ref{conclusion}.
\section{The Declarative Language}
\label{about}
LYRICS defines a TensorFlow (TF)\footnote{\url{https://www.tensorflow.org/}} environment in which learning and reasoning are integrated. LYRICS provides a short number of basic constructs, which can be used to define the problem under investigation. 

A \textit{domain} determines a collection of individuals that share the same representation space and are analyzed and manipulated in a homogeneous way. For example, a domain can collect a set of $30\times 30$ pixel images or the sentences of a book as bag-of-words. 
%The domains are then filled with their ``inhabitants'', on which the learning and reasoning will be carried on. For example, a domain called $Images$ can be defined as:
\begin{python}
	Domain(label="Images")
\end{python}
%where \textit{data\_images} is the placeholder of the input data.
%The elements in \textit{data\_images} are a sort of ``anonymous'' individuals that are collectively processed. 
%On the other hand, an \textit{individual} of a domain can also be separately specified, and a specific behavior can be defined for it.
\textit{Individuals} (i.e. elements) can be added to their domain as follows:
\begin{python}
	Individual(label="Tweety", domain=("Images"), value=img0)
\end{python}
where \textit{Tweety} is a label to uniquely identify a specific individual of the \textit{Images} domain, represented by the image \textit{img0}. This allows the user to directly reason about single individuals. The user can also provide a large amount of individuals without a specific label for each of them by specifying the tensor of their features during the domain definition.

A \textit{function} can be defined to map elements from the input domains into an element of an output domain. A unary function takes as input an element from a domain and transforms it into an element of the same or of another domain, while an $n$-ary function takes as input $n$ elements, mapping them into an output element of its output domain.
%For example, it is  possible to define arithmetic functions to operate over numbers, or encoding functions to transform elements of a domain into a latent space.
%A LYRICS function is implemented in LYRICS as a TF computational graph, taking as input a fixed number of inputs tensors representing elements of the input domains and returning an output tensor.
The following example defines a function that returns a rotated image: 
\begin{python}
	Function(label="rotate", domains=("Images"), function=RotateFunction)
\end{python}
where the FOL function is bound to its TF implementation, which in this case is the \textit{RotateFunction} function in the TF code.

A \textit{predicate} can be defined as a function, mapping elements of the input domains to truth values. %, as for example: \textit{isCat(x)}, or $f(x) > 3$.
For example, a predicate $bird$ determining whether an input patttern from the $Images$ domain contains a bird and approximated by a neural network NN is defined as:
\begin{python}
	Predicate(label="bird", domains=("Images"), function=NN)
\end{python}

It is possible to state the knowledge about the world by means of a set of  \textit{constraints}. Each constraint is a generic FOL formula using as atoms the previously defined functions and predicates. For instance, if we want to learn the previously defined predicate \textit{bird} to be invariant to rotations, the user can express this knowledge by means of the following constraint:
\begin{python}
	Constraint("forall x: bird(x) -> bird(rotate(x))")
\end{python}

Finally, any available \textit{supervision} for the functions or predicates can be directly integrated into the learning problem. LYRICS provides a specific construct where this fitting is expressed, called \textit{PointwiseConstraint}. This construct links to a computational graph where a loss is applied for each supervision. The loss defaults to the cross-entropy loss but it can be overridden to achieve a different desired behavior:
\begin{python}
	PointwiseConstraint(model, labels, inputs)
\end{python}
where \texttt{model} is a TF function like the \textit{NN} function used before fitting the supervisions \texttt{labels} on the provided  \texttt{inputs}.

\section{From Logic to Learning}

LYRICS transparently transforms a declarative description of the available knowledge applied to a set of objects into an optimization task.
In this section, we show how the optimization algorithm is derived from its declaration. %In particular, we show how single elements of the FOL language are bound to specific elements of a computational graph expressing the desired optimization process.

%\textbf{Tensorflow.} The framework is built on top of Tensorflow. This system performs computations by building a computational graph, where nodes of the graph are operations manipulating all the tensors represented by their incoming edges. TF allows automatic differentiation of a generic computational graph w.r.t. a set of Tensors (i.e. Variables) by the exploitation of the chain rule of calculus and the Backpropagation algorithm %TF provides a wide family of gradient descent algorithms to optimize the parameters.
%Since the framework implements all its components within TensorFlow, any TensorFlow model can be integrated in LYRICS.
%The proposed framework compiles a high level description of the knowledge into a computational graph by translating each piece of logic knowledge as constraints. The resulting computational graph is optimized exploiting the standard TF optimization mechanism. 
% In the following, we describe how this translation is performed.% with respect to the general architecture of LYRICS depicted in Figure~\ref{fig:architecture}.

%\begin{figure}[t]
%	\centering
%	\includegraphics[width=0.5\linewidth]{Figs/architecture.pdf}
%	\caption{An high level description of the architecture of LYRICS}
%	\label{fig:architecture}
%\end{figure}

\paragraph{\bf Domains and Individuals.} Domains of individuals allow users to provide data to the framework as tensors that represent the leaves of the computational graph. A Domain $D_i$ is always bound to a tensor $X_i \in \mathbb{R}^{d_i\times r_i}$, where $d_i$ denotes the number of individuals in the $i$-th domain and $r_i$ denotes the dimension of the representation of the data in the $i$-th domain\footnote{Here, we assume that the feature representation is given by a vector. However, the system also allows the individuals to be represented by a generic tensor.}.  Thus, individuals correspond to rows of the $X_i$ tensor.
%The tensor $X_i$ is given by the concatenation along the first dimension $X_i = [\tilde X_i, x^{(i)}_1, \dots, x^{(i)}_n]$, where  $\tilde X_i$ is the set of the anonymous individuals provided during the domain definition and $x^{(i)}_j$ is a specific individual defined by the user (i.e. an individual with a label).  
Individuals can be represented by both \textit{constant} and \textit{variable} feature tensors. By taking into account partially or totally variable features for the individuals, LYRICS allows to consider individuals as learnable objects too. 
%By allowing specific individuals to be considered variable, the user is allowed to provide the knowledge of the existence of a certain individual, even if its feature representation is unknown, and its representation will be learned to be coherent with the other pieces of knowledge provided. 
For example, given two individuals \textit{Marco} and \textit{Michelangelo} bound to a constant and a variable tensor respectively, we may want to learn the representation of \textit{Michelangelo} by exploiting some joint piece of knowledge (e.g. \texttt{fatherOf(Marco, Michelangelo) -> \texttt{similarTo(Michelangelo, Marco)}}).

\paragraph{\bf Functions and Predicates.} FOL functions allow the mapping between individuals of the input domains to an individual of the output domain, i.e. $f_i:~D^f_{i_1} \times \dots \times D^f_{i_m} \to D^f_i$, where $D^f_{i_1}, \dots, D^f_{i_m}$ are the input domains and $D^f_i$ is the output domain. On the other hand, FOL predicates allow to express the truth degree of some property for individuals of the input domains; i.e. $p_i: D^p_{i_1} \times\ldots\times D^p_{i_m} \to \{true,false\}$, where $D^p_{i_j}$ is the $j$-th domain of the $i$-th predicate. Functions and predicates are implemented using a TF architecture as explained in the previous section. If the graph does not contain any variable tensor (i.e. it is not parametric), then we say it to be \textit{given}; otherwise it will contains variables which will be automatically learned to maximize the constraints satisfaction. In this last case, we say the function/predicate to be \textit{learnable}. Learnable functions can be (deep) neural networks, kernel machines, radial basis functions, etc. 

The evaluation of a function or a predicate on a particular tuple $x_1, \dots, x_m$ of input individuals (i.e. $f_i(x_1, \dots, x_m)$ or $p_i(x_1, \dots, x_m)$) is said a \textit{grounding} for the function or for the predicate, respectively.
LYRICS, like related frameworks~\cite{diligenti2015semantic,serafini2016learning}, follows a fully grounded approach, which means that all the learning and reasoning processes take place only once functions and predicates have been fully grounded over all the possible input tuples (i.e. on the entire Cartesian product of the correponding input domains).

Let us indicate as $X_k$ the set of patterns in the domain $D_k$, then $\mathcal{X}^f_i = X^f_{i_1}  \times \dots \times X^f_{i_m}$ is the set of groundings of the $i$-th function. Similarly, $\mathcal{X}^p_i$ is the collection of groundings for the $i$-th predicate. Finally, $\mathcal{F}(\mathcal{X})= \{f_1(\mathcal{X}^f_1), f_2(\mathcal{X}^f_2), \ldots, \}$ and $\mathcal{P}(\mathcal{X})= \{p_1(\mathcal{X}^p_1), p_2(\mathcal{X}^p_2),\ldots\}$ are the outputs for all function and predicates over their corresponding groundings, respectively.

\paragraph{\bf Connectives and Quantifiers.} Connectives and quantifiers are converted using the fuzzy generalization of FOL that was first proposed by Novak~\cite{novakmathematical}. In particular, a \emph{t-norm fuzzy logic} \cite{hajek1998metamathematics} generalizes Boolean logic to variables assuming values in $[0,1]$. A t-norm fuzzy logic is defined by its t-norm that models the logical AND, and from which the other operations can be derived.
Table~\ref{tab:t-norms} shows some possible implementation of the connectives using the fundamental t-norm fuzzy logics i.e. \emph{Product, \L ukasiewicz} and \emph{G\"{o}del} logics.

\begin{table}[tb]
	\centering
	\small
	\begin{tabular}{|c|c|c|c|} % c|
		\hline
		\diagbox{op}{t-norm} & Product & Lukasiewicz  &  G\"{o}del \\ 
		\hline
		$x \land y$ & $x \cdot y$ & $\max(0, x + y -1$) & $\min(x, y$) \\
		\hline
		$x \lor y$ & $x + y - x \cdot y$ &  $\min(1, x + y)$ & $\max(x, y)$ \\
		\hline
		$\lnot x$ & $1 - x$ &  $1 - x$ & $1 - x$ \\
		\hline
		$x \Rightarrow y$ & $x\leq y?1:\frac{y}{x}$ &  $\min(1,1-x+y)$ & $x\leq y?1:y$ \\
		\hline
	\end{tabular}
	\caption{Operations performed by the units of an expression tree given the inputs $x,y$ and the used t-norm in the fundamental fuzzy logics.}
	\label{tab:t-norms}
\end{table}

In general, formulas involve more than a predicate and are evaluated on the overall grounding vectors of such predicates. The way different evaluations of a certain formula are aggregated depend on the \emph{quantifiers} occurring on its variables and their implementation. In particular, we consider the universal and existential quantifier that can be seen as a logic AND and OR applied over each grounding of the data, respectively. 

For instance, given a certain logical expression $E$ with a
universally quantified variable $x_i$ can be calculated as the average of the t-norm
generalization $t_E(\cdot)$, when grounding $x_i$ over its groundings $X_i$:
% (see \cite{diligenti2012bridging} for more details):
\begin{align}
\label{eq:forall}
\forall x_i ~ E\big(\mathcal{P}(\mathcal{X})\big) \longrightarrow \Phi_\forall(\mathcal{P}\big( \mathcal{X})\big) = \displaystyle\frac{1}{|X_i|} \sum_{x_i \in X_i}  t_E\big(\mathcal{P}(\mathcal{X})\big)
\end{align}
The truth degree of the existential quantifier is instead defined as the
maximum of the t-norm expression over the domain of the quantified variable:
\begin{align}
\label{eq:exists}
\exists x_i ~  E\big(\mathcal{P}(\mathcal{X})\big) ~~ ~\longrightarrow~ ~~ \Phi_{\exists}\big(\mathcal{P}(\mathcal{X})\big) =
\displaystyle\max_{x_i \in X_i} \; t_E\big(\mathcal{P}(\mathcal{X}) \big)
\end{align}
When multiple quantified variables are present, the conversion is recursively performed from the inner to the outer variables.
% Please note that the fuzzy formula expression is continuous and differentiable with respect to the fuzzy value of a predicate, and it can therefore easily be integrated into learning.

\paragraph{\bf Constraints.} Integration of learning and logical reasoning is achieved by translating logical expressions into continuous real-valued constraints.
The logical expressions correlate the defined elements and enforce some desired behaviour on them.
%The elements are correlated by the constraints and enforced to have some desired behavior. % Indeed, the logical formulas combine some facts expressed by the predicates by means of logical connectives with respect to variables and functions applied to them. For this reason, constraints are the main elements of this framework.

Variables, functions, predicates, logical connectives and quantifiers can all be seen as nodes of an \textit{expression tree}~\cite{diligenti2018delbp}.
The real-valued constraint is obtained by a post-fix visit of the expression tree, where the visit action builds the correspondent portion of computational graph. In particular:
\begin{itemize}
	\item visiting a \textit{variable} $x_i$ substitutes the variable with the tensor $X_i$ bound to the domain it belongs to;
	\item visiting a \textit{function} or \textit{predicate} corresponds to the grounding operation, where, first,  the Cartesian product of the input domains is computed and, then, the TF models implementing those functions are evaluated on all groundings (i.e. $f(\mathcal{X})$ or $p(\mathcal{X})$)
	\item visiting a \textit{connective} combines predicates by means of the real-valued operations associated to the connective by the considered t-norm fuzzy logic;
	\item visiting a \textit{quantifier} aggregates the outputs of the expressions obtained for the single variable groundings.
\end{itemize}

\begin{figure}
	\centering
	\includegraphics[width=\linewidth]{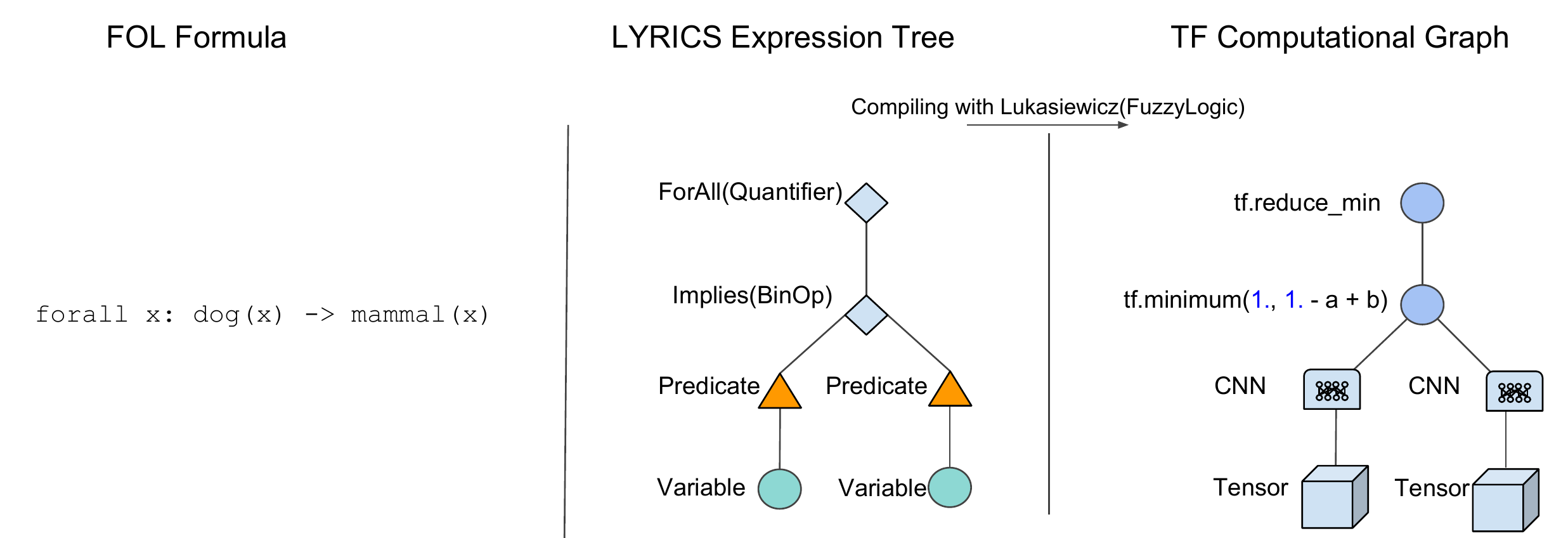}
	\caption{The translation of the FOL formula $\forall x ~ dog(x) \rightarrow mammal(x)$ into a Lyrics expression tree and then its mapping to a TF computational graph.}
	\label{fig:compilation}
\end{figure}

Figure~\ref{fig:compilation} shows the translation of a logic formula into its expression tree and successively into a TensorFlow computational graph.

It is useful for the following to consider the real-valued constraint obtained by the described compilation process of the $j$-th logical rule and implemented by a TF graph as a parametric real function $\psi_j(\mathcal{X}_j; w^i_j, w^f_j, w^p_j)$. The function $\psi_j$ takes as input the Cartesian product $\mathcal{X}_j$ of the domains of its quantified variables, returns the truth degree of the formula and it is parameterized by $w^i_j$, $w^f_j$ and $w^p_j$, which are the sets of variable tensors related to the features of learnable individuals, to the parameters of learnable functions and to the parameters of learnable predicates, respectively. Let be $W^i = \{w^i_1, w^i_2, \dots\}$, $W^f = \{w^f_1, w^f_2, \dots\}$ and $W^p = \{w^p_1, w^p_2, \dots\}$.

\paragraph{\bf Optimization Problem.} The goal of LYRICS is to build a learning process for some elements of interest (individuals, functions or predicates) by a declarative description of the desired behaviour of these elements. The desired behaviour is expressed by means of logical formulas. Thus, the optimization process is framed as finding the unknown elements which maximize the satisfaction of the set of logical formulas.  Let $\psi_j(\mathcal{X}_j; w^i_j, w^f_j, w^p_j)$ indicate the real-valued constraint related to the $j$-th formula, as previously defined. Then, the derived optimization problem is:

\begin{align*}
\max_{W^i, W^f, W^p}  \; \displaystyle \sum_{j=1}^T \lambda^j_c  \psi_j(\mathcal{X}_j; w^i_j, w^f_j, w^p_j) ,
\end{align*}
where $\lambda^j_c$ denotes the weight for the $j$-th logical constraint. These weights are considered hyper-parameters of the model and are provided by the user during constraint definition. The maximization problem can be translated into a minimization problem as follows:

\begin{align*}
\min_{W^i, W^f, W^p}  \; \displaystyle \sum_{j=1}^T \lambda^j_c  \mathcal{L}\Big(\psi_j(\mathcal{X}_j; w^i_j, w^f_j, w^p_j)\Big) ,
\end{align*}

Here, the function $\mathcal{L}$ represents any monotonically decreasing transformation of the constraints conveniently chosen according to the problem under investigation. In particular, we may exploit the following mappings:
\begin{equation}
\begin{array}{l}
\label{eq:L}
{\bf (a)}\;\;\mathcal{L}\Big(\psi_j(\mathcal{X}_j; w^i_j, w^f_j, w^p_j) \Big)=1-\psi_j(\mathcal{X}_j; w^i_j, w^f_j, w^p_j) \ ,\\
{\bf (b)}\;\;\mathcal{L}\Big(\psi_j(\mathcal{X}_j; w^i_j, w^f_j, w^p_j) \Big)=-\log\Big(\psi_j(\mathcal{X}_j; w^i_j, w^f_j, w^p_j)\Big) \ .
\end{array}
\end{equation}
These specific choices for the function $\mathcal{L}$ are directly related to the \L ukasiewicz and Product t-norms, indeed they are \emph{additive generators} for these t-norm fuzzy logics. For more details on generated t-norms we recommend e.g. \cite{klement2004triangular2}.

\section{Learning and Reasoning with Lyrics}
\label{zoo}
This section presents a list of examples illustrating the range of learning tasks that can be expressed in the proposed framework. In particular, it is shown how it is possible to force label coherence in semi-supervised or transductive learning tasks, how to implement collective classification over the test set and how to perform model checking.  Moreover, we applied the proposed framework to two standard benchmarks: document classification in citation networks and term chunking in natural language text. The examples are presented using LYRICS syntax directly to show that the final implementation of a problem fairly retraces its abstract definition. 
%This is a precise design choice which we believed fundamental for a programming framework intended for users from the wide AI community.
The software of the framework and the experiments are made available at "provided as supplemental materials".
%\url{https://github.com/GiuseppeMarra/lyrics}
% In this overview of AI toy problems, we range between classic \emph{Machine Learning} techniques relying on data and inference--driven processes purely based on constraints satisfaction (e.g. in section \ref{queen}).
% In particular, in LYRICS we are able to express any FOL constraint also including arithmetical operators as special primitives. In addition, we can exploit the TensorFlow tools to model more complex learning settings, i.e.  a generative task.

\begin{figure}[th!]
	\centering
	\vspace{-0.2cm}
	\subfigure[]{\label{fig:semisupervised_a}\includegraphics[width=0.47\linewidth]{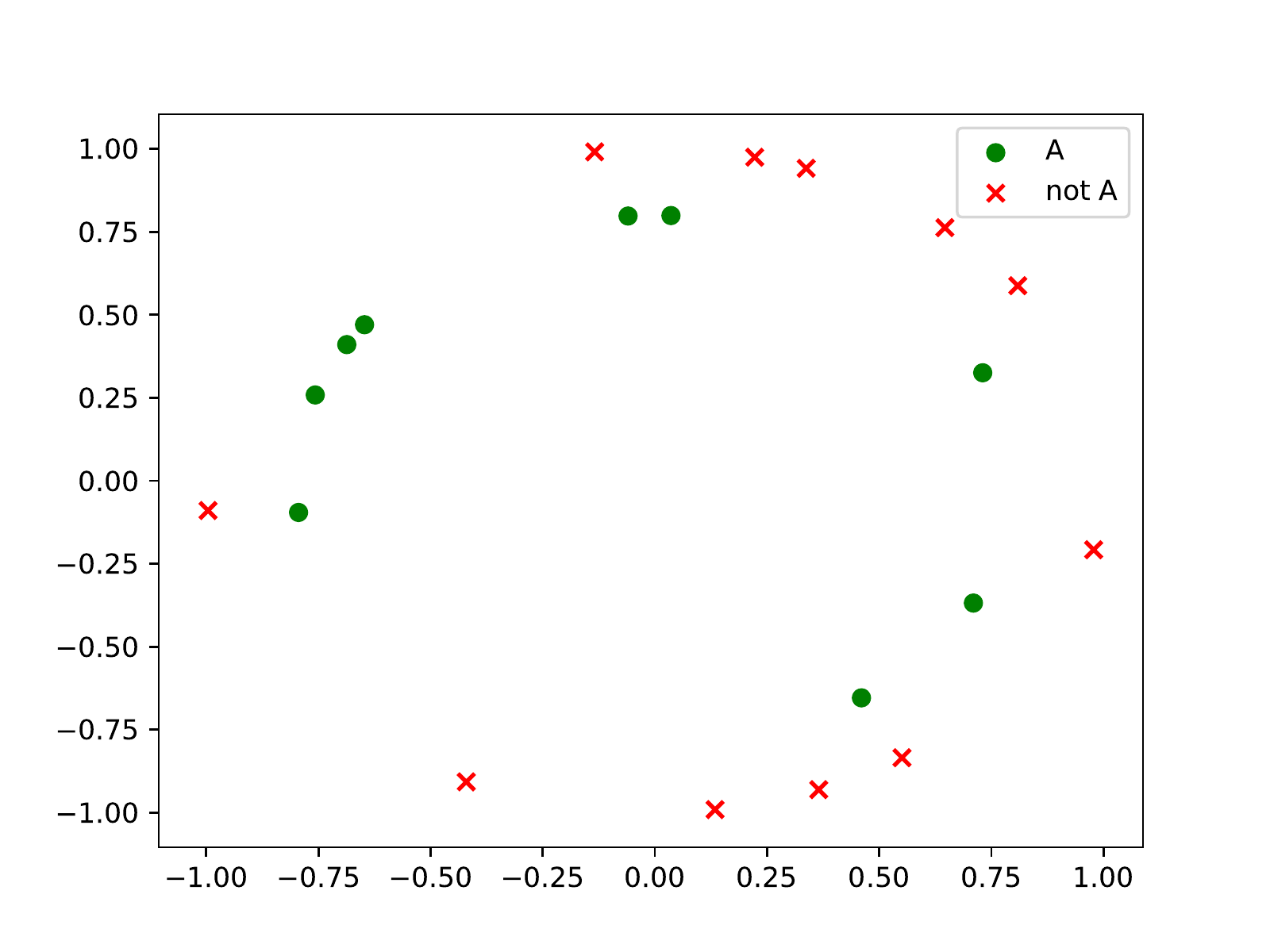}}
	~
	\subfigure[]{\label{fig:semisupervised_b}\includegraphics[width=0.47\linewidth]{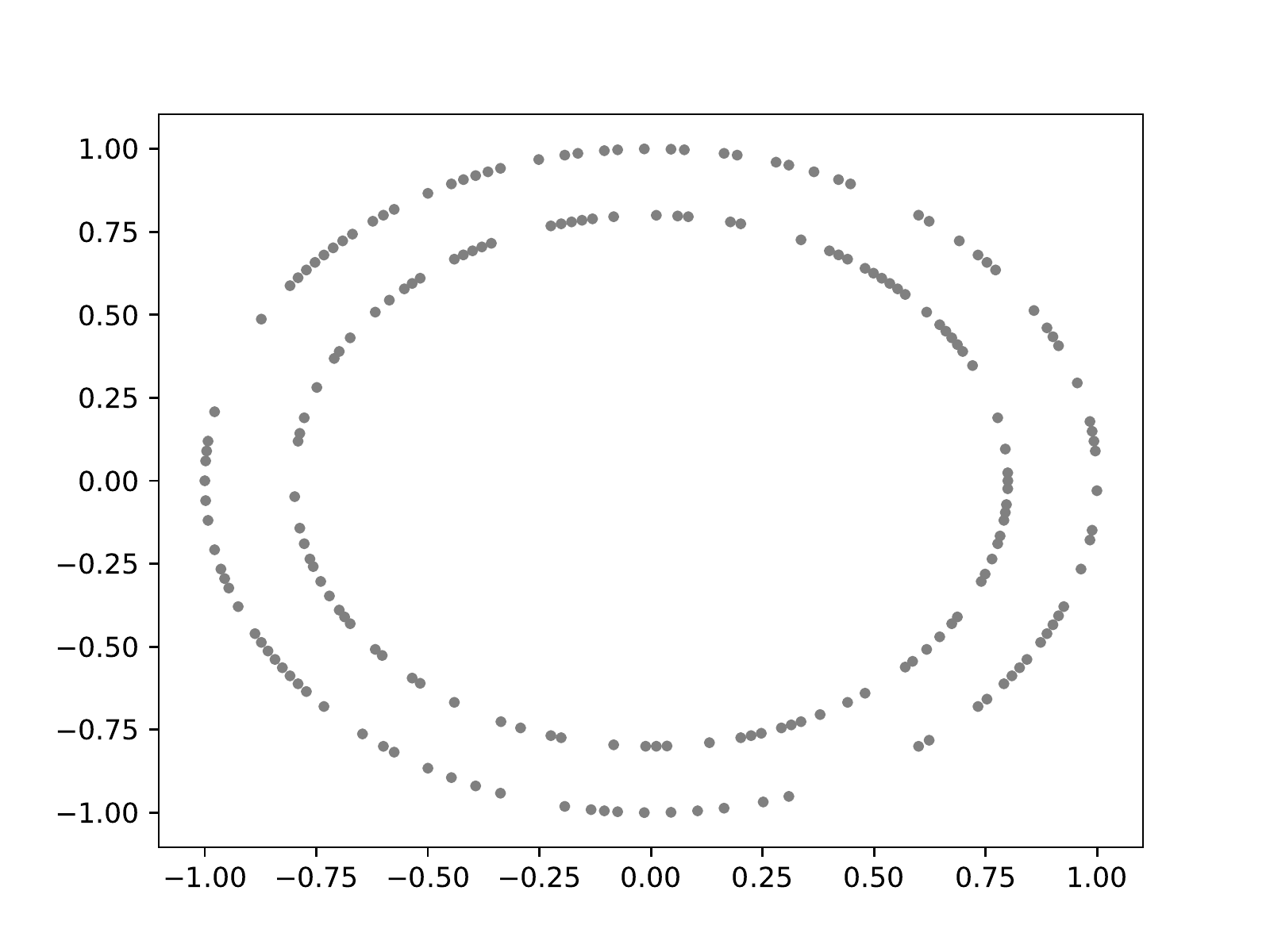}}
	\vspace{-0.2cm}
	\subfigure[]{\label{fig:semisupervised_c}\includegraphics[width=0.47\linewidth]{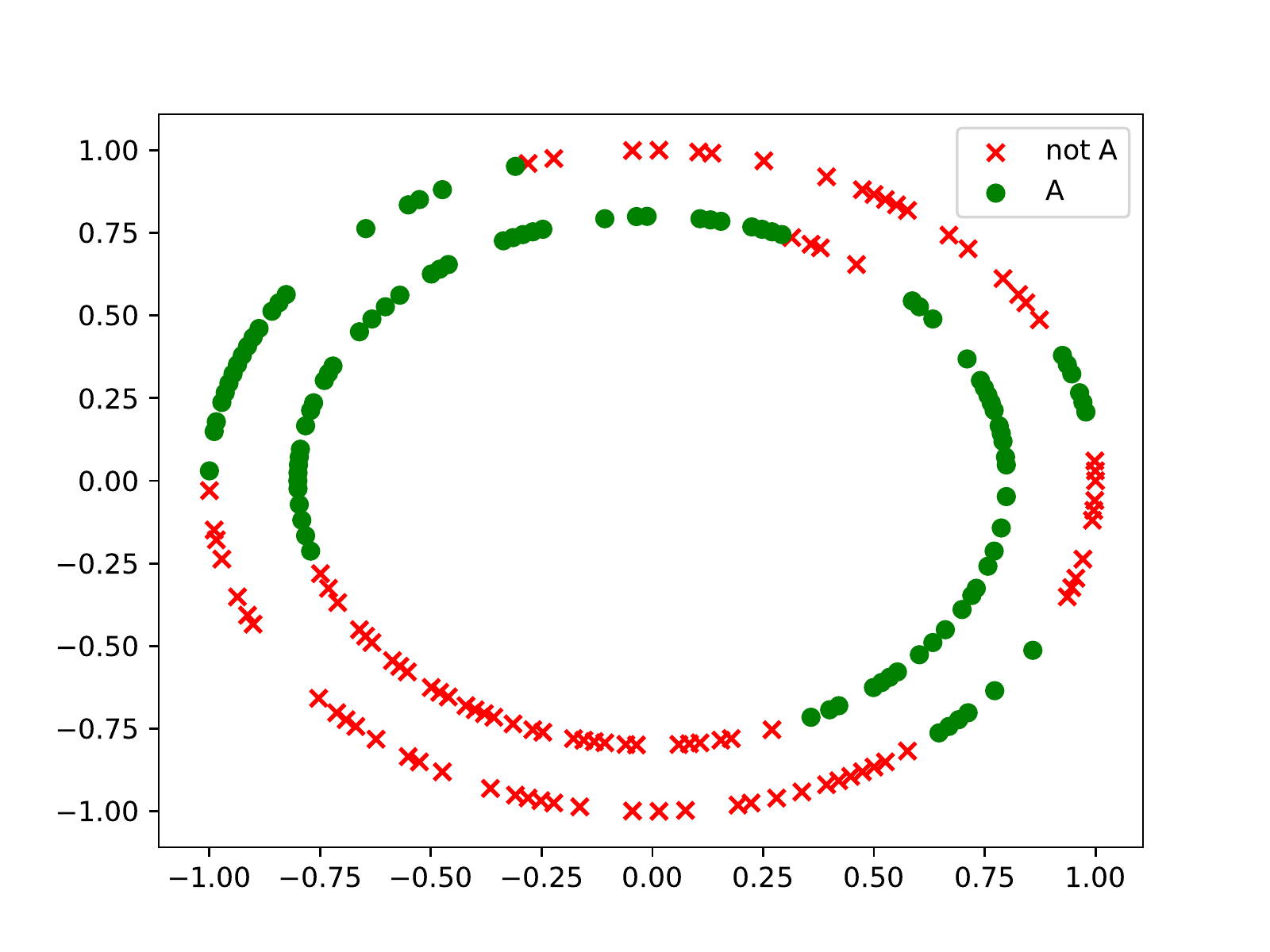}}
	~
	\subfigure[]{\label{fig:semisupervised_d}\includegraphics[width=0.47\linewidth]{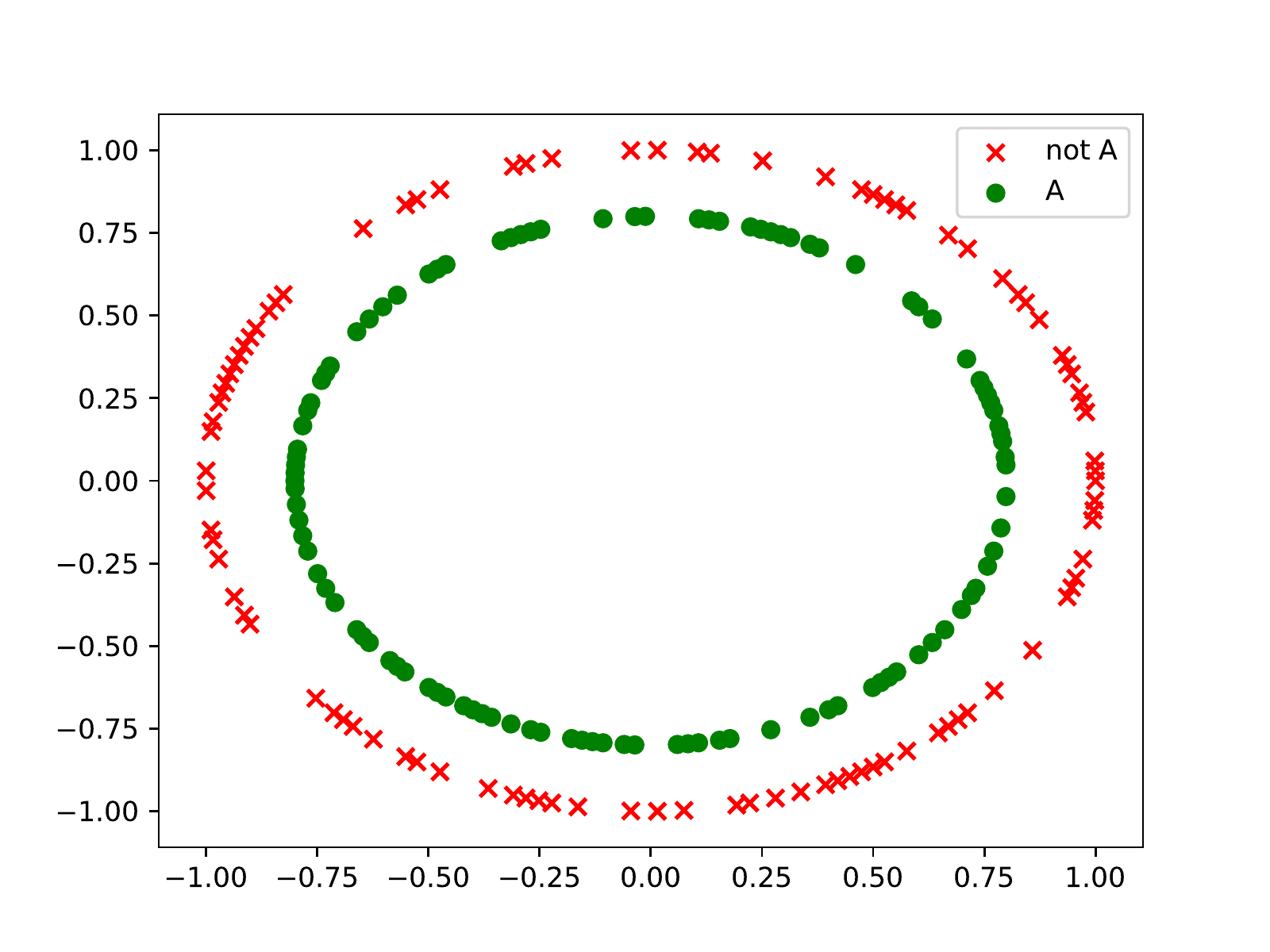}}
	\caption{Semi-supervised Learning: (a) data that is provided with a positive and negative supervision for class $A$; (b) the unsupervised data provided to the learner; (c) class assignments using only the supervised examples; (d) class assignments using learning from examples and constraints.}
	\label{fig:semisupervised}
\end{figure}

%%%%%%%%%%%%%%%%%%%%%%%%%%%%%%%%%%%%%%%%%%%%%%%%%
\paragraph{Semi-Supervised Learning. }
In this task we assume to have available a set of $420$ points distributed along an outer and inner circle. The inner and outer points belong and do not belong to some given class $A$, respectively. A random selection of $20$ points is supervised (either positively or negatively), as shown in Figure~\ref{fig:semisupervised_a}. The remaining points are split into $200$ unsupervised training points, shown in Figure~\ref{fig:semisupervised_b} and $200$ points left as test set. A neural 
network is assumed to have been created in TF to approximate the predicate $A$.
The network can be trained by making it fit the supervised data. So, given the vector of data \texttt{X}, a neural network \texttt{NN\_A} and the vector of supervised data \texttt{X\_{s}}, with the vector of associated labels \texttt{y\_{s}}, the supervised training of the network can be expressed by the following:
\begin{python}
	# Definition of the data points domain.
	Domain(label="Points", data=X)
	# Approximating the predicate A via a NN.
	Predicate("A", ("Points"), NN_A)
	# Fit the supervisions
	PointwiseConstraint(A, y_s, X_s)
\end{python}
%where \texttt{PointwiseConstraint(predicate, targets, examples)} is a special constraint forcing the predicate to fit of the supervised data using a cross-entropy loss.

Let's now assume that we want to express manifold regularization for the learned function: e.g. points that are close should be similarly classified. This can be expressed as:
\begin{python}
	# Predicate stating whether 2 patterns are close.
	Predicate("Close", ("Points","Points"), f_close)
	# Manifold regularization constraint.
	Constraint("forall p:forall q: Close(p,q)->(A(p)<->A(q))")
\end{python}
where \texttt{f\_close} is a given predicate determining if two patterns are close.
The training is then re-executed starting from the same initial conditions as in the supervised-only case.

Figure~\ref{fig:semisupervised_c} shows the class assignments of the patterns in the test set, when using only classical learning from supervised examples. Finally, Figure~\ref{fig:semisupervised_d} presents the assignments when learning from examples and constraints.

\begin{figure}[t]
	\centering
	\subfigure[]{\label{fig:collective_a}\includegraphics[width=0.315\linewidth]{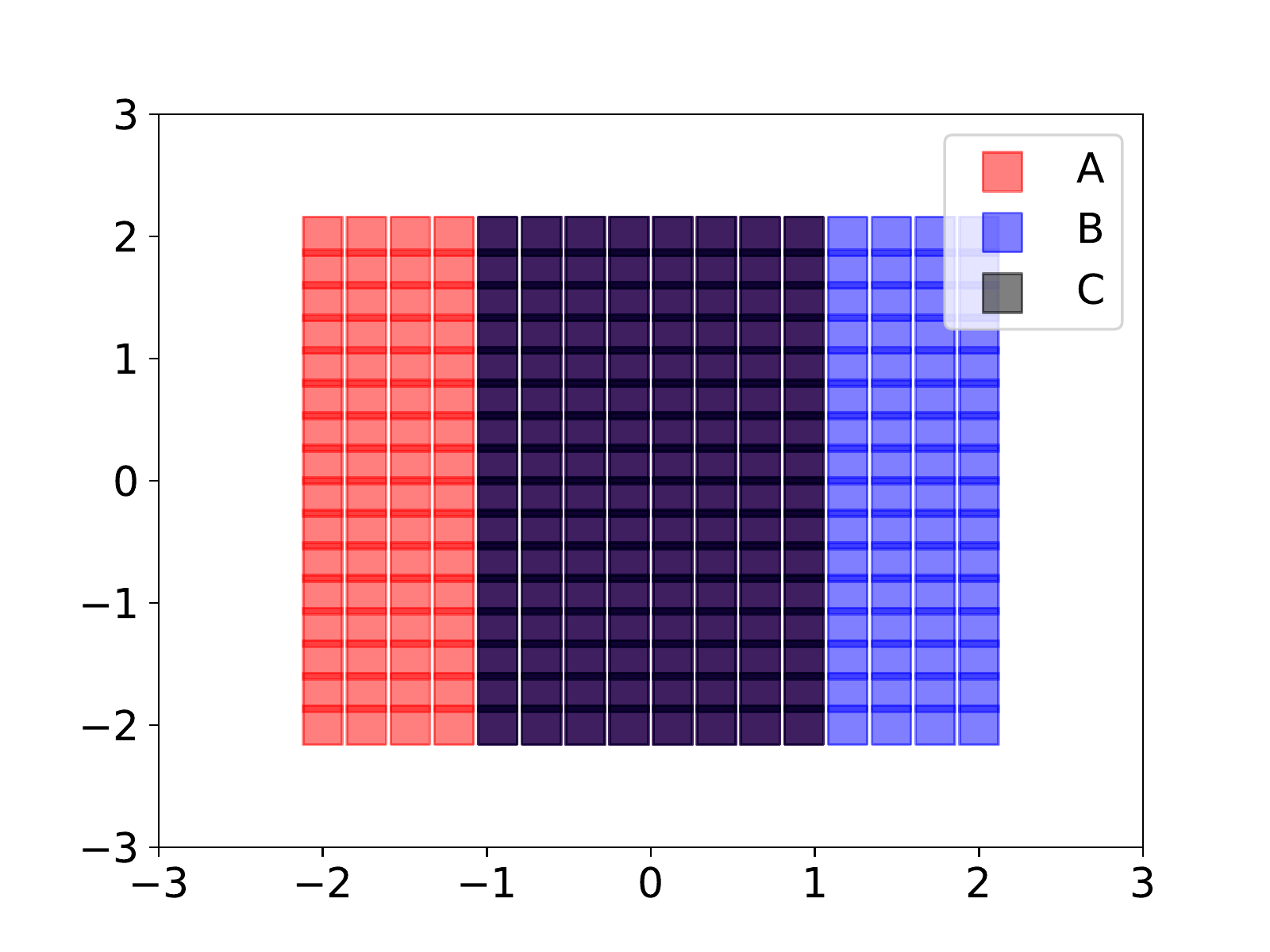}}
	~
	\subfigure[]{\label{fig:collective_b}\includegraphics[width=0.315\linewidth]{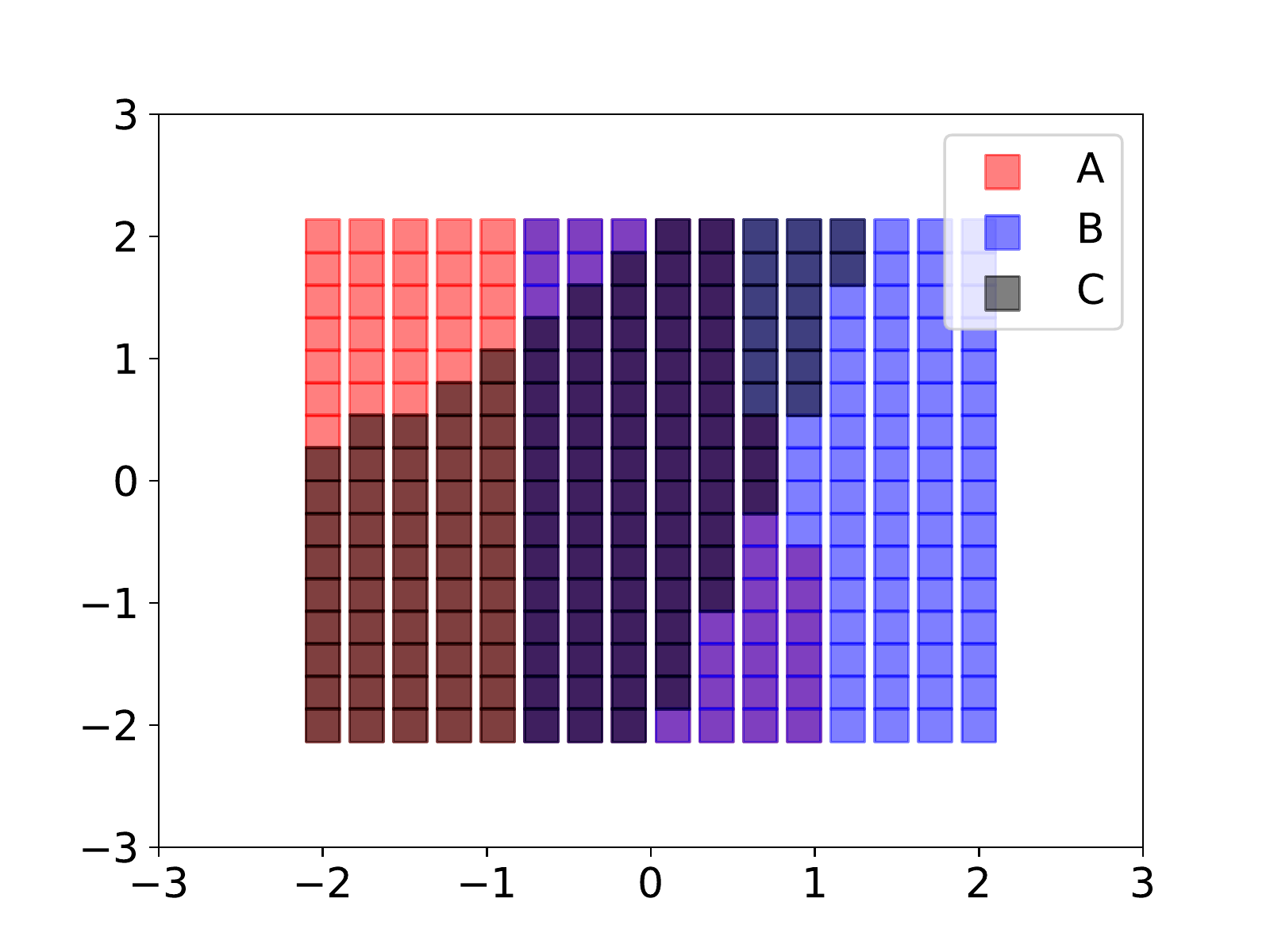}}
	\label{fig:semisupervised_data}
	\subfigure[]{\label{fig:collective_c}\includegraphics[width=0.315\linewidth]{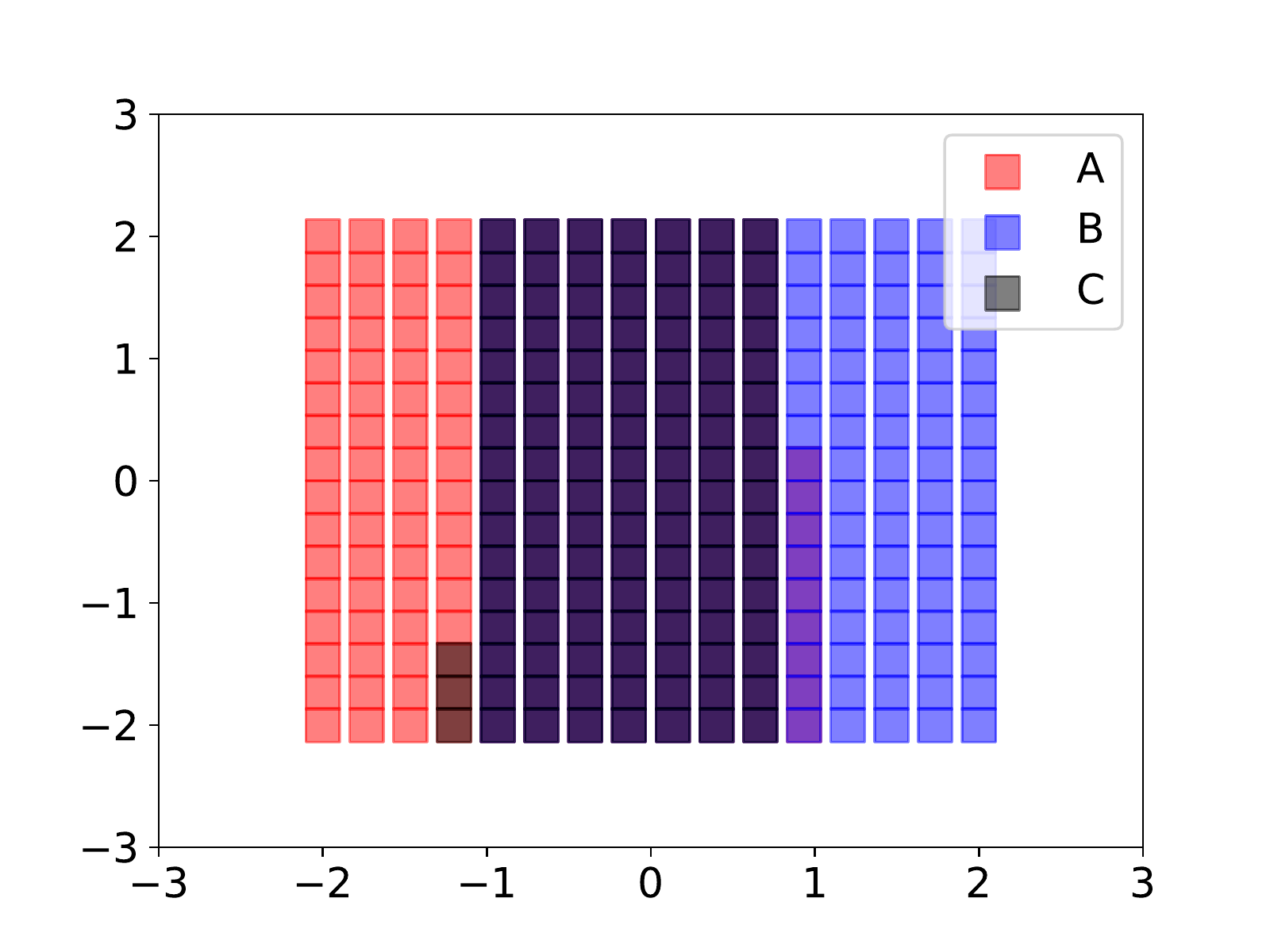}}
	~
	\caption{Collective classification: (a) classes assignments; (b) the predictions after the supervised step; (c) the predictions with collective classification and rules satisfaction (best viewed in colors).}
	\label{fig:collective}
\end{figure}

%%%%%%%%%%%%%%%%%%%%%%%%%%%%%%%%%%%%%%%%%%%%%%%%%
\paragraph{Collective Classification. }
Collective classification~\cite{sen2008collective} performs the class assignments exploiting any known correlation among the test patterns. This paragraph shows how to exploit these correlations in LYRICS. 
Here, we assume that the patterns are represented as $\mathbb{R}^2$ datapoints. The classification task is a multi-label problem where the patterns belongs to three classes $A, B, C$. In particular, the class assignments are defined by the following membership regions:
${\bf A}=[-2,1]\times[-2,2], {\bf B}=[-1,2]\times[-2,2],{\bf C}=[-1,1]\times[-2,2]$.
These regions correspond to three overlapping rectangles as shown in Figure~\ref{fig:collective_a}.
The examples are partially labeled and drawn from a uniform distribution on both the positive and negative regions for all the classes.

In a first stage, the classifiers for the three classes are trained in a supervised fashion using  a two-layer neural network taking four positive and four negative examples for each class. This is implemented via the following declaration:
\begin{python}
	Domain(label="Points", data=X)
	Predicate(label="A",domains=("Points"),NN_A)
	Predicate(label="B",domains=("Points"),NN_B)
	Predicate(label="C",domains=("Points"),NN_C)
	PointwiseConstraint(NN_A, y_A,  X_A)
	PointwiseConstraint(NN_B, y_B,  X_B)
	PointwiseConstraint(NN_C, y_C,  X_C)
\end{python}

The test set is composed by $256$ random points and the assignments performed by the classifiers after the training are reported in Figure~\ref{fig:collective_b}. %The F1 score for $A, B$ and $C$ is 0.95, 0.95 and 0.72, respectively.
In a second stage, it is assumed that it is available some prior knowledge about the task at hand. In particular, any pattern must belongs to (at least) one of the classes $A$ or $B$. Furthermore, it is known that class $C$ is defined as the intersection of $A$ and $B$.
%The classifiers trained in the first stage provide some initial predictions for each pattern. 
The collective classification step is performed by seeking the class assignments that are as close as possible to the initial classifier predictions but also respect the logical constraints on the test set:
\begin{python}
	Constraint("forall x: A(x) or B(x)")
	Constraint("forall x:(A(x) and B(x)) <-> C(x)")
	# Minimize the distance from prior values
	PointwiseConstraint(A, priorsA,  X_test)
	PointwiseConstraint(B, priorsB,  X_test)
	PointwiseConstraint(C, priorsC,  X_test)
\end{python}
where \texttt{X\_test} is the set of test datapoints and \texttt{priorsA}, \texttt{priorsB}, \texttt{priorsC} denote the predictions of the classifiers to which the final assignments have to stay close.
As we can see from Fig.\ref{fig:collective_c}, the collective step fixes some wrong predictions. % and the F1 score for $A, B$ and $C$ reaches $0.98$, $1$ and $0.92$, respectively.

%It is also possible to mark a set of constraints as test only, in order to perform model checking. Model checking can be used as a fundamental step to perform rule deduction using the Inductive Logic Programming techniques~\cite{muggleton1994inductive}.

\paragraph{Model checking. }
In this example, we show how the framework can be used to perform model checking. Let us consider a simple multi-label classification task where the patterns belong to two classes $A$ and $B$, and $B$ is contained in $A$. This case models a simple hierarchical classification task. In particular, the classes are defined by the following membership regions:
${\bf A}=[-2,2]\times[-2,2],\quad{\bf B}=[-1,1]\times[-1,1]$.
A set of points \texttt{X} is drawn from a uniform distribution in the $[-3,3]\! \times\![-3,3] $ region.
Two neural network classifiers are trained to classify the points using the vectors of supervisions $y_A$ and $y_B$ for the predicates $A$ and $B$, respectively:
\begin{python}
Domain(label="Points", data=X)
Predicate(label="A", domains=("Points"),NN_A)
Predicate(label="B", domains=("Points"),NN_B)
PointwiseConstraint(NN_A, y_A,  X)
PointwiseConstraint(NN_B, y_B,  X)
\end{python}
It could be interesting to check if some given rule has been learned by the classifiers. To this hand, LYRICS allows to mark a set of constraints as test only, in order to perform model checking. In this case, constraints are only used to compute the degree of satisfaction of the corresponding FOL formulas over the data. For example, we checked the degree of satistaction of all possible formulas in Disjunctive Normal Form (DNF) that are universally quantified with a single variable. Only the constraint:
\begin{python}
 Constraint("forall x: (not A(x) and not B(x)) or (A(x) and not B(x)) or (A(x) and B(x))")
\end{python}
has a high truth degree (0.9997). As one could expect, the only fully-satisfied constraint (translated from DNF to its minimal form) is indeed $\forall x B(x) \rightarrow A(x)$, that states the inclusion of $B$ in $A$. Model checking can be used as a fundamental step to perform rule deduction using the Inductive Logic Programming techniques~\cite{muggleton1994inductive}.

\paragraph{Chunking. }
\label{sec:chunking}
Given a sequence of words, term chunking (or shallow parsing) is a sequence tagging task aiming at linking constituent parts of sentences (nouns, verbs, adjectives, etc.) into phrases that form a single semantic unit.
Following the seminal work by Collobert et al.~\cite{collobert2011natural}, many papers have applied deep neural networks to text chunking. In this paper, deep learner is used to learn from examples as in classical supervised learning. Then we perform collective classification to fix some misclassification made by the network, according to certain logical rules expressing available prior knowledge.

We used the CoNLL 2000 shared task dataset~\cite{tjong2000introduction} to test the proposed methodology. The dataset contains $8936$ training and $893$ test English sentences. The task uses $12$ different chunk types, which correspond to $22$ chunk labels when considering the position modifiers. In particular, some labels have a $B$ and $I$ modifier to indicate for beginning and intermediate position in the chunk, respectively. For example, $BVP$ indicates the start of a verbal phrase and $IVP$ an intermediate term of the verbal phrase.
The final performance is measured in terms of F1-score, computed by the public available script provided by the shared task organizers.

We selected the classifier proposed by Huang et al.~\cite{huang2015bidirectional} as our baseline, which is one of the best performers on this task. We used a variable portion of training phrases from the training set, ranging from $5$\% to $100$\%, to train the classifier, reusing the same parameters reported by the authors. The trained networks have then been applied on the test set providing an output score for each label for each term.
It is well known that the output of the trained networks may not respect the semantic consistencies of the labels. For example, an intermediate token for a label must follow either a begin or intermediate one for the same label.
For example, $\forall~x~\forall~t~BNP(x,t) \Rightarrow \lnot IVP(x,t+1) \land \lnot IPP(x,t+1) \land \lnot IADVP(x,t+1) \land \ldots$ expresses that if the $t$-th token is marked as the begin of a nominal phrase $BNP$ the following token can not be an intermediate verbal $IVP$, intermediate prepositional $IPP$ or intermediate adverbial $IADVP$ phrase.
A small sample of the constraints stating the output consistency can be expressed in FOL using the following statements:
\[
\small
\begin{array}{l}
\forall~x~\forall~t~BNP(x,t) \Rightarrow \lnot IVP(x,t+1) \land \lnot IPP(x,t+1) \land \lnot IADVP(x,t+1) \land \ldots\\
\forall~x~\forall~t~BVP(x,t) \Rightarrow \lnot INP(x,t+1) \land \lnot IPP(x,t+1) \land \lnot IADVP(x,t+1) \land \ldots\\
\forall~x~\forall~t~BPP(x,t) \Rightarrow \lnot IVP(x,t+1) \land \lnot INP(x,t+1) \land \lnot IADVP(x,t+1) \land \ldots\\
\forall~x~\forall~t~INP(x,t) \Rightarrow [\lnot IVP(x,t+1) \land \lnot IPP(x,t+1) \land \lnot IADVP(x,t+1) \land \ldots\\
\forall~x~\forall~t~IVP(x,t) \Rightarrow [\lnot INP(x,t+1) \land \lnot IPP(x,t+1) \land \lnot IADVP(x,t+1) \land \ldots\\
\forall~x~\forall~t~IPP(x,t) \Rightarrow [\lnot IVP(x,t+1) \land \lnot INP(x,t+1) \land \lnot IADVP(x,t+1) \land \ldots\\
\forall~x~\forall~t~INP(x,t+1) \Rightarrow BNP(x,t) \lor INP(x,t)\\
\forall~x~\forall~t~IVP(x,t+1) \Rightarrow BVP(x,t) \lor IVP(x,t)\\
\forall~x~\forall~t~IPP(x,t+1) \Rightarrow BPP(x,t) \lor IPP(x,t)\\
\ldots
\end{array}
\] 
where $P(x,t)$ indicates the output of the network associated to label $P$ for the phrase $x$ and the $t$-th term in the phrase.

In order to evaluate the proposed methodology, collective classification is performed to assign the labels in order to minimize the distance from the network outputs, acting as priors, while maximizing the verification of the constraints built from the previously reported rules.
Table~\ref{tab:chunking} reports the F1 results for the different percentages of supervised phrases used to train the network. The results have been evaluated both on all classes, and then zooming in for some of the rare classes that are often wrongly classified. The effect of the rules is overall mildly positive as most of the tags can be correctly predicted by the supervised examples. However, the effect of the knowledge is more clear when zooming in to see the effect on the some of the less common tags ($ADJ, ADV, PRT, SBAR$): since not enough examples are observed for these tags, the extra knowledge allows to improve their classification. Since these tags are relatively rare the overall effect on the metrics is not large on this dataset, but it is a very promising start to allow the application of pos tagging to challenging domains.

%DA FARE
\begin{table}[t]
	\centering
	\begin{tabular}{llccccc}
	    &&\multicolumn{5}{c}{\% data in training set}\\
		&& 5 & 10 & 30 & 50 & 100 \\
		\hline
		\multirow{2}{*}{F1}& NN & 87.39 & 89.55 & 92.15 & 93.31 & 94.18 \\
		&LYRICS  &  87.75 &   89.78 &  92.26 &   93.53 &  94.27 \\
		%LYRICS\_cons\_exist\_n & 87.48 & 89.71 & 92.23 & 93.48 & \bfseries  94.27 \\
		\hline
		\multirow{2}{*}{F1(rare tags)~~~~}&NN & 56.24 & 60.84 & 75.19 & 76.74 & 79.42 \\
		&LYRICS ~~~~& 57.65 & 61.36 &  75.68 &   77.45 & 79.71 \\
		%LYRICS\_cons\_exist\_n & \bfseries 58.24 &\bfseries 63.8 & 74.95& 76.84 & \bfseries 79.9 \\
		\hline
	\end{tabular}
	\caption{CoNLL2000 evaluation script on all the classes and on the less common pos tags that have an initial lower performance.}
	\label{tab:chunking}
\end{table}

%%%%%%%%%%%%%%%%%%%%%%%%%%%%%%%%%%%%%%%%%%%%%%%%%

\paragraph{Document Classification on the Citeseer dataset. }
This section applies the proposed framework to a standard ML dataset.
The CiteSeer dataset\footnote{\url{https://linqs.soe.ucsc.edu/data}}~\cite{lu2003link} consists of $3312$ scientific papers, each one assigned to one of $6$ classes: Agents, AI, DB, ML and HCI. The papers are not independent as they are connected by a citation network with $4732$ links. Each paper in the dataset is described via its bag-of-word representation, which is a vector having the same size of the vocabulary with the $i$-th element having a value equal to $1$ or $0$, depending on whether the $i$-th word in the vocabulary is present or not present in the document, respectively. The dictionary consists of $3703$ unique words. This learning task is expressed as:
\begin{python}
	Domain(label="Papers", data=X)
	Predicate("Agents",("Papers"), Slice(NN, 0))
	Predicate("AI",("Papers"), Slice(NN, 1))
	Predicate("DB",("Papers"), Slice(NN, 2))
	Predicate("IR",("Papers"), Slice(NN, 3))
	Predicate("ML",("Papers"), Slice(NN, 4))
	Predicate("HCI",("Papers"), Slice(NN, 5))
\end{python}
where the first line defines the domain of scientific articles to classify, and one predicate for each class is defined and bound to an output of a neural network $NN$, which features a softmax activation function on the output layer.

The domain knowledge that if a paper cites another one, they are likely to share the same topic, is expressed as:
\begin{python}
	Predicate("Cite",("Papers","Papers"),f_cite)
	Constraint("forall x: forall y: Agent(x) and Cite(x, y) -> Agent(y)")
	Constraint("forall x: forall y: AI(x) and Cite(x, y) -> AI(y)")
	Constraint("forall x: forall y: DB(x) and Cite(x, y) -> DB(y)")
	Constraint("forall x: forall y: IR(x) and Cite(x, y) -> IR(y)")
	Constraint("forall x: forall y: ML(x) and Cite(x, y) -> ML(y)")
	Constraint("forall x: forall y: HCI(x) and Cite(x, y) -> HCI(y)")
\end{python}
where \texttt{f\_cite} is a given function determining whether a pattern cites another one. Finally, the supervision on a variable size training set can be provided by means of:
\begin{python}
	PointwiseConstraint(NN, y_s, X_s)
\end{python}
where \texttt{X\_s} is a subset of the domain of papers where we enforce supervisions \texttt{y\_s}.

\begin{table}[t]
	\centering
	\begin{tabular}{lccccc}
	\toprule
		& \multicolumn{5}{c}{\footnotesize{\% data in training set}}\\
		& 10 & 30 & 50 & 70 & 90 \\
    \midrule
		NN & 60.08 & 68.61& 69.81& 71.93 &72.59 \\
		LYRICS & \textbf{67.39}& \textbf{72.96} & \textbf{75.97}& \textbf{76.86} &\textbf{78.03}\\
	\bottomrule
    \end{tabular}
	\caption{Citeseer dataset: comparison of the 10-fold average accuracy obtained by supervised training of a neural network (NN), and by learning the same NN from supervision and logic knowledge in LYRICS for a variable percentage of training data. Bold values indicate statistically significant improvements.}
	\label{tab:citeseer}
\end{table}
\begin{table}[t]
	\centering
	\begin{tabular}{lc}
	    \toprule
	    Method & Accuracy\\
		\midrule
		Naive Bayes & 74.87 \\
		ICA Naive Bayes &76.83\\
		GS Naive Bayes &76.80\\
		Logistic Regression &73.21\\
		ICA Logistic Regression &77.32\\
		GS Logistic Regression &76.99\\
		Loopy Belief Propagation & 77.59\\
		Mean Field &77.32 \\
		NN & 72.59 \\
		LYRICS & \textbf{78.03}\\
		\bottomrule
	\end{tabular}
	\caption{Citeseer dataset: comparison of the 10-fold average accuracy obtained by content based and network based classifiers and by learning from supervision and logic knowledge in LYRICS.}
	\label{tab:citeseer2}
\end{table}
Table~\ref{tab:citeseer} reports the accuracy obtained by a neural network with one hidden layer (200 hidden neurons) trained in a supervised fashion and by training the same network from supervision and logic knowledge in LYRICS, varying the amount of available training data and averaged over 10 random splits of the training and test data. The improvements over the baseline are statistically significant for all the tested configurations.
Table~\ref{tab:citeseer2} compares the neural network classifiers against other two content-based classifiers, namely logistic regression (LR) and Naive Bayes (NB), and against collective classification approaches using network data: Iterative Classification Algorithm (ICA)~\cite{neville2000iterative} and Gibbs Sampling (GS)~\cite{lu2003link} both applied on top of the output of LR and NB content-based classifiers.
Furthermore, the results against the two top performers on this task: Loopy Belief Propagation (LBP)~\cite{sen2008collective} and Relaxation Labeling through Mean-Field Approach (MF)~\cite{sen2008collective} are reported. The accuracy values are obtained as average over 10-folds created by random splits of size $90$\% and $10$\% of the overall data for the train and test sets, respectively.
Unlike the other network based approaches that only be run at test-time (collective classification), LYRICS can distill the knowledge in the weights of the neural network. The accuracy results are the highest among all the tested methodologies in spite that the underlying neural network classifier trained only via the supervisions did perform slightly worse than the other content-based competitors.

\section{Conclusions}
\label{conclusion}
This paper presents a novel and general framework, called LYRICS, to bridge logic reasoning and deep learning.
The framework is directly implemented in TensorFlow, allowing a seaming-less integration that is architecture agnostic. The frontend of the framework is a declarative language based on First--Order Logic. Throughout the paper are presented a set of examples illustrating the generality and expressivity of the framework, which can be applied to a large range of tasks. 

Future developments of the proposed framework include a learning mechanism of the weights of the constraints. This would allow to consider more general rule schemata that will be weighted with coefficients automatically learned by the parameter optimization according to the degree of satisfaction of any rule. This will improve the framework especially to deal with soft constraints expressing some statistical co-occurrence among the classes involved in the learning problem. Moreover, the differentiability of fuzzy logic could suggest new methods for learning a set of constraints in logical form that may be understandable.

% Future developments of the proposed framework can explore a learning mechanism of the weights of the constraints, which would allow to safely insert rules without the certainty of their validity. Moreover, the differentiability of fuzzy logic could open the doors to new methods for logical rule learning.

 \bibliographystyle{splncs04}
 \bibliography{references}

\end{document}